\pdfoutput=1

\documentclass[11pt]{article}

\usepackage{acl}

\usepackage{times}
\usepackage{graphicx}
\usepackage{soul}
\usepackage{parskip}
\usepackage{times}
\usepackage{latexsym}
\usepackage{booktabs}
\usepackage{multirow}
\usepackage{enumitem}

\usepackage{stfloats}

\usepackage[T1]{fontenc}

\usepackage[utf8]{inputenc}

\usepackage{microtype}

\usepackage{algorithmic}
\usepackage[ruled,vlined]{algorithm2e}
\SetKwFunction{function}{\textbf{function}}

\usepackage{booktabs}

\title{Improving Zero-Shot Event Extraction via Sentence Simplification}

\author{Sneha Mehta \\
 Twitter Cortex \\
  \texttt{snehamehta@twitter.com} \\\And
  Huzefa Rangwala \\
  George Mason University  \\
  \texttt{rangwala@gmu.edu}  \\\And 
   Naren Ramakrishnan \\
  Virginia  Tech \\
  \texttt{naren@cs.vt.edu} \\}

\begin{document}
\maketitle
\vspace{-4em}

\begin{abstract}
The success of sites such as ACLED and Our World in Data have demonstrated the massive utility of extracting events in structured formats from large volumes of textual data in the form
of news, social media, blogs and discussion forums.
Event extraction can provide a window into ongoing geopolitical crises and yield actionable intelligence. 
With the proliferation of large pretrained language models, Machine Reading Comprehension (MRC) has emerged as a new paradigm for event extraction in recent times. In this approach, event argument extraction is framed as an extractive question-answering task. One of the key advantages of the MRC-based approach is its ability to perform zero-shot extraction. However, the problem of long-range dependencies, i.e., large lexical distance between trigger and argument words and the difficulty of processing syntactically complex sentences plague MRC-based approaches.
%
%
In this paper, we present a general approach to improve the performance of MRC-based event extraction by performing unsupervised sentence simplification guided by the MRC model itself. We evaluate our approach on the ICEWS geopolitical event extraction dataset, with specific attention to `Actor' and `Target' argument roles. We show how such context simplification can improve the performance of MRC-based event
extraction by more than 5\% for actor extraction and more than 10\% for 
target extraction. 

\end{abstract}

\section{Introduction}\label{sec:intro}

\begin{figure}[ht!]
 \centering
  \includegraphics[width=0.5\textwidth]{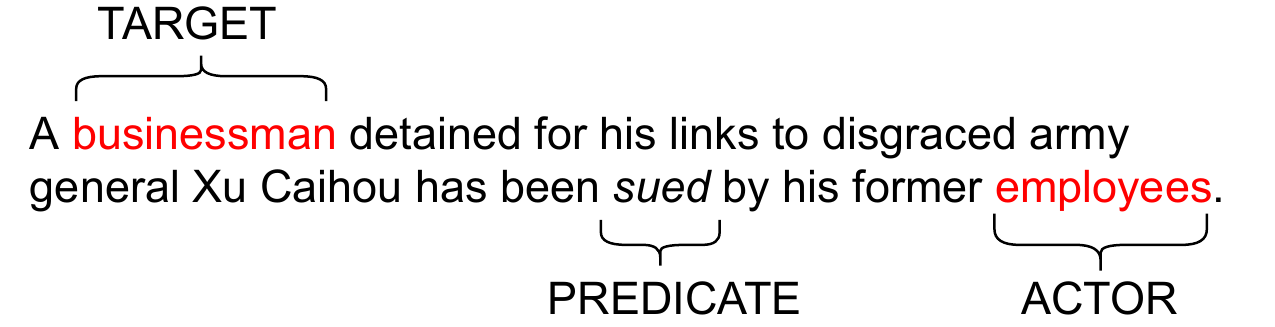}
 \caption{An example of an event of the type `Bring lawsuit against' from the ICEWS dataset.}
 \label{fig:example}
\vspace{-1.3em}
\end{figure}

With the proliferation of social media, microblogs and online news, we are able to gain a real-time understanding of events happening around the world. By ingesting large unstructured datasets and converting them into structured formats such as (actor, event, target) tuples we can make rapid progress in 
systems for event forecasting ~\citep{embers}, real-time event coding ~\citep{Saraf:2016:EAA:2939672.2939737} or other applications that can grant organizations a strategic advantage. 
Historically, this has been enabled by efforts such as ICEWS\footnote{https://dataverse.harvard.edu/dataverse/icews} \& GDELT\footnote{https://www.gdeltproject.org/}. 
These systems rely on event extraction technology to populate their knowledge bases. Fig.~\ref{fig:example} gives an example of an event `Bring lawsuit against' from the ICEWS dataset. Extraction involves identifying entities (businessman, employees) corresponding to argument roles `Actor' and `Target'.
The \emph{event}  is triggered by the predicate `sued' in the figure. However, the extraction technology employed by these
systems relies on pattern-based approaches that use handcrafted patterns designed to extract entities and events~\citep{Boschee2013}. Even though pattern-based methods have high precision, they fail to work on unseen event types and with new event categories. 
Hence, there is a need to explore extraction methods that can extend beyond a fixed domain.
Modern approaches for event extraction~\citep{chen2015event,nguyen-etal-2016-joint,wadden-etal-2019-entity} rely on fine-grained annotations and suffer from data scarcity issues and error propagation due to pipeline systems.


With the success of large scale pretrained language models on machine reading comprehension (MRC) tasks~\citep{devlin2018bert,liu2019roberta,huang-etal-2018-zero}, a new paradigm for event extraction based on MRC has surfaced~\citep{du-cardie-2020-event,liu-etal-2020-event}. 
In this approach, event argument extraction is posed as a span extraction problem from a context conditioned on a question for each argument. This approach is promising because it mitigates some of the issues faced by traditional approaches, such as relying on upstream systems to extract entities/triggers and hence sidestepping the error propagation problem in pipeline systems. It also gives rise to the possibility of zero-shot event extraction and hence the ability to extend to new domains which is traditionally hard due to difficulties in collecting high-quality labeled training data. However, MRC models struggle with long-range dependencies and syntactic complexities.  For instance,  \citet{liu-etal-2020-event} observe that one
typical error from their MRC-based extraction system is related to long-range dependency between an argument and a trigger,
accounting for 23.4\% errors on the ACE-2005 event dataset~\cite{doddington-etal-2004-automatic} (here ``long-range'' denotes that
the distance between a trigger and an argument is greater than or equal to 10 words).~\citet{du-cardie-2020-event} observe that one of the failure modes of their extraction system is sentences with complex sentence structures containing multiple
clauses, each with trigger and arguments. These observations make a promising case for complexity reduction or context simplification for MRC systems. 

To mitigate the above problem and to reduce the syntactic complexity we propose an unsupervised approach that is guided by a scoring function that incorporates syntactic fluency, simplicity and the confidence of an MRC model(\S~\ref{sec:methods}) Our key contributions are:
\begin{enumerate}
    \item The exploration of sentence and context simplification to help mitigate thelong-range dependency problem for MRC based zero-shot event extraction (\S~\ref{sec:ch3_res}). 
    \item Experimental results on the ICEWS political event datasets including a detailed analysis of areas of selective superiority of our approach.
    \item Followup analysis to demonstrate how 
    simplification can be controlled based on desired factors such as preference for high performance for a certain argument role.
\end{enumerate}


\begin{figure*}[ht!]
 \centering
  \includegraphics[width=0.98\textwidth]{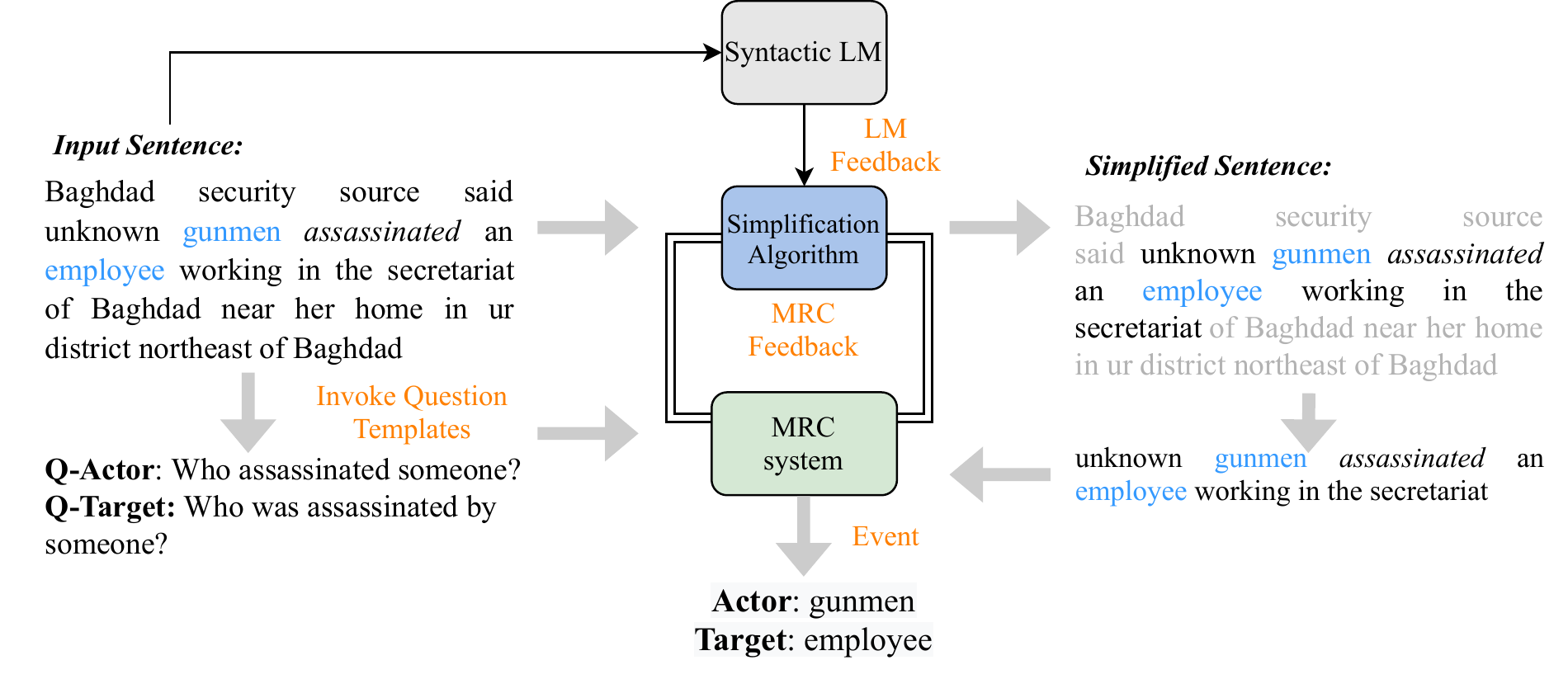}
 \caption{The RUSS sentence simplification approach.}
 \label{fig:disamb_overview}
\end{figure*}



\section{Methodology}\label{sec:methods}


Reading comprehension models can be brittle to subtle changes in context. They can be thrown-off by syntactic complexity, especially when the questions are not specific and do not include words overlapping with the context. Moreover, long range dependencies between the trigger/predicate and the argument can also throw the model off. For this purpose, we propose an M\textbf{R}C-guided \textbf{U}nsupervised \textbf{S}entence \textbf{S}implification algorithm (RUSS), that iteratively performs deletions and extractions from the context in search for a higher-scoring candidate. The score function incorporates components that ensure sentence fluency, information preservation and the confidence of the target MRC model.  Fig.~\ref{fig:disamb_overview} gives an overview of the proposed approach.

Concretely, given that an event has been detected in a sentence, the task is to identify the arguments of the detected event. For instance, in Fig.~\ref{fig:example} the task is to identify the arguments `Actor' and `Target' of the event ‘Bring lawsuit against’. The algorithm takes the QA pairs corresponding to each argument role as input. The QA generation procedure for the dataset used in this paper for evaluation is outlined in Appendix~\ref{sec:qa_gen}. 
Table~\ref{tab:example-record} shows the generated QA-pair for the arguments Actor and Target for the event shown in Fig.~\ref{fig:example}.

\begin{table}[ht!]
\small
\centering
\caption{An example of a generated QA record for an event shown in Fig.~\ref{fig:example} from the ICEWS dataset. The highlighted words are answers to the generated questions.}
\label{tab:example-record}
\begin{tabular}{l|p{5cm}}
\toprule
\textbf{Sentence} & A {\color{red}businessman} detained for his links to disgraced army general Xu Caihou has been \textit{sued} by his former {\color{red}employees}. \\
\textbf{Q-Actor} & Who \textit{sued} someone? \\
\textbf{Q-Target} & Who was \textit{sued} by someone? \\
\bottomrule

\end{tabular}
\end{table}

\subsection{Sentence Simplification Algorithm}\label{sec:ctx_disamb}

Given an input sentence $s$ and a list of questions $\{q_1,...,q_n\}$ corresponding to different arguments, our algorithm iteratively performs two operations on the sentence -- deletion and extraction, in search for a higher-scoring sentence and outputs a candidate simplification $c$. For generating candidates, the algorithm first obtains the constituency parse tree of the context using a span-based constituency parser~\citep{joshi-etal-2018-extending}. It then sequentially performs two operations on the parse tree -- deletion and extraction.

\paragraph{Deletion}
In this operation, the algorithm sequentially drops subtrees from the parse tree corresponding to different phrases. Note that the subtrees with the NP (Noun-Phrase) label are omitted because it is expected that many entities that form event arguments will be noun phrases and deleting them from the sentence would result in significant information loss. 

\paragraph{Extraction}
This operation simply extracts a phrase, specifically corresponding to the the S and SBAR labels as the candidate sentence. This allows us to select different clauses in a sentence and remove remaining peripheral information.

These operations generate multiple candidates.  Candidates with fewer than a threshold of $t$ words are filtered out. We heuristically determine $t=5$.
From the remaining candidates, a highest-scoring candidate is chosen based on the score function described in the next section(\S~\ref{sec:score_fun}). The algorithm terminates if the maximum score assigned to a candidate in the current iteration does not exceed the previous maximum score. The simplification algorithm RUSS is outlined as Algorithm~\ref{algo:RUSS} and the candidate generation algorithm is outlined as Algorithm~\ref{algo:cand_gen}.

\begin{algorithm}[h!]
 \caption{Sentence Simplification Algorithm -- RUSS}
 \label{algo:RUSS}
\footnotesize
\SetAlgoLined
\KwIn{$sentence:=s$, $questions=\{q_1,..q_n\}$}
\KwOut{$simplification:=c$}
\textbf{Function} \texttt{RUSS($s$)}: \\
 $maxIter \leftarrow M$\\
\For{$iter \in maxIter$}{
 $candidates \leftarrow \texttt{generateCandidates}(c)$ \\
 $scores \leftarrow \emptyset$ \\ 
 $maxScore \leftarrow 0$ \\
 \For{$cand \in candidates$}{
 $scores \leftarrow scores \cup \nu_{lm}^a*\nu_{entity}^b*\nu_{pred}^c*\prod\nu_{rc_{role_i}^{r_i}}$ \\
 }
 $currMax \leftarrow max(scores)$ \\
 \If{$currMax > maxScore$}{
   $maxScore \leftarrow currMax$ \\
    $c \leftarrow candidates[argmax(scores)]$
    
    }
 }
 \KwRet $c$
\end{algorithm}

\begin{algorithm}[h!]
 \caption{Candidate Generation Algorithm}
 \label{algo:cand_gen}
\footnotesize
\SetAlgoLined
\KwIn{$sentence:=s$}
\KwOut{$candidates$}
\textbf{Function} \texttt{generateCandidates($s$)}:\\
$parseTree \leftarrow getParseTree(s)$ \\ 
$toRemove \leftarrow \emptyset$ \\
$extractions \leftarrow \emptyset$ \\
$candidates \leftarrow \emptyset$ \\
$phraseTags \leftarrow getValidPhraseTags()$ \\
 \For{$pos \in parseTree.positions$}{
 \If{$parseTree[pos] \in phraseTags$}{
   $toRemove \leftarrow toRemove \cup parseTree[pos].leaves$
    }
    
 \If{$pos.label \in [S, SBAR]$}{
   $extractions \leftarrow extractions \cup parseTree[pos].leaves$
    }
 }
 \For{$phrase \in toRemove$}{
 $candidate \leftarrow s.replace(phrase, \emptyset)$ \\
  \If{$candidate.length > t$}{
     $candidates \leftarrow candidates \cup candidate$
    }
 }
  \For{$phrase \in extractions$}{
  \If{$phrase.length > t$}{
     $candidates \leftarrow candidates \cup candidate$
    }
}    
 \KwRet $candidates$
\end{algorithm}

\subsection{Scoring Function}\label{sec:score_fun}
We score a candidate as a product of different scores corresponding to fluency, simplicity and its amenability to the downstream MRC model.

\paragraph{LM Score ($\nu_{lm})$}
This score is designed to measure the language fluency and structural simplicity of a candidate sentence. Instead of using LM-perplexity we use the syntactic log-odds ratio (SLOR)~\citep{pauls-klein-2012-large,carroll-etal-1999-simplifying} score to measure the fluency. SLOR was also shown to be effective in simplification to enhance text readability~\citep{kann-etal-2018-sentence,kumar2020iterative}. Given a trained language model (LM) and a sentence $s$, SLOR is defined as
\begin{equation}
\small
    SLOR(s) = \frac{1}{|s|}(ln(P_{LM}(s)) - ln(P_{U}(s))
\end{equation}
where $P_{LM}$ is the sentence probability given by the language model, $P_{U}(s) = \prod_{w\in s}P(w)$
is the
product of the unigram probability of a word $w$ in
the sentence, and $|s|$ is the sentence length.
SLOR essentially penalizes a plain LM’s probability by unigram likelihood and the length. It
ensures that the fluency score of a sentence is not
penalized by the presence of rare words.
A probabilistic language model (LM) is often
used as an estimate of sentence fluency. In our work, instead of using a plain LM we use a syntax-aware LM, i.e., in addition to words, we use part-of-speech (POS) and dependency tags as inputs to the LM ~\cite{zhao2018language}. For a word $w_i$
, the input to the syntax-aware LM is $[e(w_i); p(w_i); d(w_i)]$, where $e(w_i)$ is
the word embedding, $p(w_i)$ is the POS tag embedding, and $d(w_i)$ is the dependency tag embedding.
Note that our LM is trained on the original train corpus.
Thus, the syntax-aware LM helps to identify candidates that are structurally ungrammatical.

\paragraph{Entity Score ($\nu_{entity})$} Entities help identify the key information of a sentence and therefore are also useful in measuring meaning preservation. The desired argument roles are also entities. Thus, if any entity detected in the original sentence is omitted from a candidate the entity score for that candidate is 0, else it is set to 1.

\paragraph{Predicate Score ($\nu_{pred})$}
This score preserves the event predicates in a candidate. It checks if a candidate contains any predicate of interest corresponding to the event detected (Table~\ref{tab:icews_predicates}). If it does not then $\nu_{pred}$ is set to 0, else it is set to 1.

\paragraph{MRC Score ($\nu_{rc})$} 
Transformer-based MRC models can be brittle to subtle changes in context. To make the context robust to the MRC model this score allows us to control the complexity of context with respect to the confidence of the MRC model. It is computed separately for each role. Each argument of an event is a span in the context. $\nu_{rc_{role_i}^{r_i}}$ is the score of the best span in the context for the argument role $i$, where the score of a candidate span is defined as $S·T_x + E·T_y$ where $S \in R^H$ is a start vector and $E \in R^H$ is an end vector as defined in \citet{devlin-etal-2019-bert}. $T_x$ and $T_y$ are the final layer representations from the BERT model of the $x^{th}$ and $y^{th}$ tokens in the context. Note that for a valid span,  $y > x$. This score is computed separately for each argument role (Actor and Target in Example~\ref{fig:example}). The  importance of the $i^{th}$ role can be controlled by the exponent $r_i$. The total contribution of each role is computed as the product of score corresponding to each role, given by $\prod\nu^{r_i}_{rc_{role_i}}$.

The final score of a candidate $c$ is computed as follows: 
\begin{equation}
\small
    \nu(c) = \nu_{lm}(c)^{a}*\nu_{entity}^{b}(c)*\nu_{pred}^{c}(c)*\prod\nu^{r_i}_{rc_{role_i}}(c)
\end{equation}

Note that $b, c$ can be either $1$ or $0$ since $\nu_{entity}$ and $\nu_{pred}$ are binary. In later sections, we evaluate how the simplification can be controlled by varying the constants $r_i$'s.

\section{Datasets and Metrics}
We evaluate RUSS on the ICEWS event dataset\footnote{https://dataverse.harvard.edu/dataverse/icews} from years 2013 to 2015. In this dataset, event data consists of coded interactions between socio-political actors (i.e., cooperative or hostile actions between individuals, groups, sectors and nation states) mapped to the CAMEO~\footnote{https://parusanalytics.com/eventdata/data.dir/cameo.html} ontology.
These events are in the form of triples consisting of a source actor, an event type (according to the CAMEO taxonomy of events), and a target actor. For evaluation, we aim to extract the Actor and Target roles for event types shown in Table ~\ref{tab:icews_predicates}. We use the data from years 2013-2015 for training and 2016 for testing/evaluation in section~\ref{sec:sup}. See Appendix for distribution of event types.

\subsection{Evaluation}
We evaluate the performance of an MRC system before and after simplification in the cross-domain zero-shot setting. In this setting, we emulate a no-resource scenario, i.e. using the MRC system out-of-the-box in a target domain. We do not finetune a pretrained MRC model with the generated QA dataset. Rather, the aim is to assess the model performance in a zero-shot setting, without using any training data from the target domain whatsoever. We used the pretrained BERT model finetuned on the SQUAD 2.0 dataset~\citep{rajpurkar-etal-2016-squad} and use the predictor API provided here \footnote{\url{https://docs.allennlp.org/models/v2.4.0/models/rc/predictors/transformer_qa/}}. 
 We further conduct follow-up analysis to study the controllability of simplification by performing ablation analysis and assessing model performance for different values of score component coefficients. For evaluation, we extracted the best span(s) and computed an exact match F1 score ~\cite{Seo2017BidirectionalAF} matching the span against the ground truth answer.

\section{Results \& Discussion}\label{sec:ch3_res}
 




\begin{table*}[ht!]
\small
\centering
\caption{Results of zero-shot event extraction on the ICEWS dataset. $\nu_{lm}$ coefficient $a=1.5$ and $\nu_{entity}$ coefficient $b=1$ for all settings in which simplification is performed. $\Delta$ $+ve$ indicates the \% of records for which F1 improves after simplification, $\Delta$ $-ve$ indicates the \% of records for which F1 becomes worse after simplification and $\Delta$ same indicates the \% of records for which F1 remains unchanged.}
\label{tab:zero-shot-results}
\begin{tabular}{l|l|llll|llll}
\toprule
& &         \multicolumn{4}{c|}{\textbf{Actor}}       & \multicolumn{4}{c}{\textbf{Target}}              \\ 
&  \textbf{Method}      & F1 & $\Delta$ $+ve$ & $\Delta$ $-ve$ & $\Delta$ same & F1 & $\Delta$ $+ve$ & $\Delta$ $-ve$ & $\Delta$ same \\ \midrule
1 & No simplification & 0.412    & -              & -              & -             & 0.354    & -              & -              & -             \\
2 & $c=0,r_1=1,r_2=1$ & 0.431    & \textbf{10.99}\%        & 6.54 \%        & 82.45\%       & 0.391    & \textbf{17.35}\%        & 7.9\%          & 74.9\%        \\
3 & $c=1,r_1=0,r_2=0$ & 0.429    & 10.81\%        & 6.57 \%        & 82.61\%       & 0.390    & 16.54\%        & 7.53\%          & 75.93\%     \\
4 & $c=1,r_1=1,r_2=1$ & 0.424    & 10.5\%         & 6.3 \%         & 83.1\%        & 0.387    & 16.29\%        & 7.64\%         & 76.05 \%      \\
5 & $c=1,r_1=3,r_2=0$ & \textbf{0.435 }   & 9.72\%         & \textbf{5.67}\%         & \textbf{84.6}\%        & 0.391    & 16.89\%        & 7.97\%         & 75.12\%       \\
6 & $c=1,r_1=0,r_2=3$ & 0.427    & 10.54\%        & 6.95\%         & 82.5\%        & \textbf{0.391}    & 16.12\%        & \textbf{7.29}\%         & \textbf{76.59}\%       \\ \bottomrule
\end{tabular}
\end{table*}

The results of zero-shot extraction on the ICEWS dataset are outlined in Table ~\ref{tab:zero-shot-results}. 
In the baselines used, simplification is performed with score function exponents for $\nu_{lm}$ as $a=1.5$ and $\nu_{entity}$ as $b=1$ held constant while varying $c$ for $\nu_{pred}$, $r_1$ for $\nu_{actor}$ and $r_2$ for $\nu_{target}$.
With no simplification we get F1 scores of 0.412 and 0.354 for actor and target roles respectively.
For the most basic setting for simplification with $c=0$, $r_1=1$ and $r_2=1$ scores improve by 4.6\% for actor prediction to 0.431 and by 10.4\% to 0.391 for target prediction respectively which shows that simplifying context can further improve a powerful model like BERT in a cross-domain zero-shot setting. For actor prediction, out of 37,894 records we find that for 10.99\% records, F1 score improves after simplification, for 6.54\% records F1 decreased after simplification and for the rest the score remained unchanged. For target prediction, for 17.4\% records scores improve where as for 7.9\% records the scores decreased and for the rest of the records, the scores remained unchanged. 
After introducing the predicate score ($c=1$) we see that these improvements drop slightly. This is counter-intuitive, because one would expect model performance to improve when relevant predicates are present in the context. We attribute this behavior to the MRC model leveraging the language priors in the training data to predict the answers. For instance, the model could predict the subject of the predicate as an answer for `Who' type of questions.

Next, we increase the coefficients of Actor and Target roles from 1 to 3. The reason why we choose an odd number for this exponent is because sometimes for bad candidates the RC scores can be negative and since all the scores are combined in a multiplicative way, raising a negative score to an even power would reverse the desired effect. Observing the results in rows 5 \& 6 of Table~\ref{tab:zero-shot-results} we can see that percentage of sentences with similar scores before and after simplification have increased. We can also observe that percentage of sentences for which scores decrease after simplification have also decreased. We can conclude that by raising the coefficients of role specific scores we can make the simplification models more robust to inaccurate simplifications for those roles. We also observe, when $r_1=3$, we get the highest F1 for actor prediction, an improvement of 5.6\% over no simplification and for $r_2=3$ we can an F1 on-par with the highest obtained in row 2. Our results clearly indicate the benefit of simplification over no simplification and also the gradual improvement in scores when the argument coefficients $r_1,r_2$ are varied from 0 to 3.

\subsection{Long Range Dependencies}
Mean length of the original sentences is 32 words where as mean length of the sentences after simplification is 22 words (row 2 setting). This indicates that simplification doesn't make sentences too short as is intuitive because cutting relevant information would harm the performance.

We proceeded to investigate if simplification has addressed the long-range dependency problem. We look at statistics concerning the distance between the predicate and its arguments (Actor and Target) for the setting $c=0,r_1=1,r_2=1$, that is, when the predicate score($\nu_{pred}$) is not taken into account. As Table~\ref{tab:zero-shot-results} indicates for $11\%$ of the records performance increases after simplification. We find that for those records the average distance between the predicate and its argument Actor is about 13 words and the average distance between the predicate and target in the simplified context is about 10 words. For the argument Target the average distance between the predicate and target is about 8 words for original and about 6 words for the simplified context. 

We see that RUSS cuts about 3 words for Actor prediction and 2 words for Target prediction on average. We conclude that a certain percentage of improvement comes from cutting down the distance between the predicates and arguments hence mitigating the long-range dependency problem.

Moreover, we observe that actor prediction performance is better than that of target prediction. This could be attributed to the fact that for sentences that are in active construction, the subject of the predicate is a candidate for the actor and in active constructions the distance between the predicate and its subject will be small. However, for a complex sentence a clause or a phrase can occur between the subject and the predicate. In this case, the distance between them increases and hence we expect the performance to go down. We quantitatively verify this as follows -- the average length between the actor and predicate  in the candidate simplification  for which the scores improve is about 7 words and for which the scores decreases is about 8 words. Hence, whenever the distance between the predicate and actor is large the performance tends to become worse confirming our hypothesis.


 
\subsection{Qualitative Analysis}
Table ~\ref{tab:qualitative_pos} lists some cases in which simplification helps MRC system perform better.
In the first example, the proposed method deleted the word `personally' from the original sentence (\textbf{Sentence}) to obtain the simplified sentence (\textbf{Simplified}) as shown in the Table. The question posed to RC model was ``Who is being apologized to by someone'' and the ground truth answer is ``the opposition''. For the original context the model extracts ``Nawaz Sharif'' as the answer which is the wrong, whereas after removing the adverb ``personally'', it gets the correct answer. Note, that this decreases the distance between the predicate \textit{apologized} from its argument Nawaz Sharif. In the second example, RC model extracts the closest noun phrase ``Xu Caihou'' as answer which is incorrect. Simplification deletes the prepositional phrase ``to disgraced army general Xu Caihou'' aiding the RC model in extracting the correct answer. Note, that in this case it was especially important to delete the above phrase due to the inherent ambiguity of construction. This case also highlights the limitations of the current RC systems as the system was not able to successfully associate employees with businessman and predicted the noun-phrase closest to the predicate \textit{sued}. In the third example, there was segmentation error in the ICEWS dataset and two sentences were strung together as seen in the Table. RUSS successfully deleted the unrelated sentence aiding the RC system in extracting the correct answer.

\begin{table*}[!ht]
\centering
\caption{Qualitative examples of zero-shot performance of RC model before and after simplifying the context using the proposed algorithm. Underlined words are ground truth answers, emphasized words are predicates(triggers) and strikethrough indicates that words were removed by the algorithm.}
\footnotesize
\label{tab:qualitative_pos}
\begin{tabular}{lp{13cm}}
\toprule
\textbf{Question} & Who is being apologized to by someone?\\
\textbf{Sentence} & Islamabad prime minister Nawaz Sharif  personally \textit{apologized} to \underline{{\color{blue}the opposition}} today for what  he called unfortunate comments made  against PPP's Aitzaz Ahsan \\
\textbf{Answer} & Nawaz Sharif \\
\textbf{{\color{brown}Simplified}} & Islamabad prime minister Nawaz Sharif \st{personally} \textit{apologized} to \underline{{\color{blue}the opposition}} today for what  he called unfortunate comments made  against PPP's Aitzaz Ahsan \\
\textbf{Answer} & the opposition \\ 
\midrule
\textbf{Question} & Who is being sued by someone? \\
\textbf{Sentence} & Scmp a \underline{{\color{blue}businessman}} detained for his links to disgraced army general Xu Caihou has been \emph{sued} by his former employees \\
\textbf{Answer} & Xu Caihou \\
\textbf{{\color{brown}Simplified}} & \st{Scmp a} \underline{{\color{blue}businessman}} detained for his links \st{to disgraced army general Xu Caihou} has been \emph{sued} by his former employee \\
\textbf{Answer} & businessman \\
\midrule
\textbf{Question} & Who is being accused of something? \\
\textbf{Sentence} & Thus after having attacked the two elected to his party ump Brice Hortefeux and Claude Goasguen it was accused of pressure and insults. Rachida Dati has \emph{accused} \underline{{\color{blue}Claude Goasguen}} to take to her because she had refused to sleep with him and this during an altercation proved by the Canard Enchan. \\
\textbf{Answer} & Rachida Dati \\
\textbf{{\color{brown}Simplified}} & \st{Thus after having attacked the two elected to his party ump Brice Hortefeux and Claude Goasguen it was} \st{accused of pressure and insults.} Rachida Dati has \emph{accused} \underline{{\color{blue}Claude Goasguen}} \st{to take to her} because she had refused to sleep with him and this during an altercation proved by the Canard Enchan.\\
\textbf{Answer} & Claude Goasguen \\
\bottomrule
\end{tabular}
\end{table*}

\subsection{Error Analysis}
From 6.54\% records for which the score decreased after simplification (row 2 of Table~\ref{tab:zero-shot-results}) for 39.5\% records from those the prediction from the original context is a substring of the prediction from the simplified context. This means that for some cases, both the original and the simplified context facilitate the correct answer, rather it is the case that the answer from the simplified context contains extra information for which it is penalized during F1 score computation. For example consider the context ``baghdad security source said unknown gunmen assassinated an employee working in the secretariat of baghdad near her home in ur district northeast of baghdad" which after running the simplification algorithm is shortened to ``\st{in baghdad security source said} unknown gunmen assassinated an employee working in the secretariat \st{of baghdad near her home in ur district northeast of baghdad}". (The strikethrough text represents the text deleted by the proposed algorithm.) For the question; ``Who was assassinated by someone?" when presented with the original context the RC model extracts ``an employee" whereas after removing the strikethrough text, RC model extracts ``an employee working in the secretariat". The ground truth answer for this is ``employee". As can be seen both answers are correct but the simplified contex is penalized for extra words. Interestingly, such cases make up 48\% of records for cases for which performance improves after simplification. This is intuitive, because since context becomes shorter and more precise after simplification and hence one expects RC models to extract more precise answers. The fact that this happens in 39.5\% cases in the reverse scenario is surprising.

\subsection{In-Domain Training}\label{sec:sup}
In sections~\ref{sec:ch3_res}.1-~\ref{sec:ch3_res}.3 we saw how RUSS improved zero-shot event extraction performance in the cross-domain setting. In this section, we consider the scenario when we have labeled in-domain training data available and we wish to investigate if simplification can help improve performance when the MRC system has been finetuned on in-domain data. We use the BERT-base-cased model~\cite{devlin-etal-2019-bert} as our base model and finetune it on the ICEWS train dataset. We finetune all layers as opposed to just the classification layer as we observe large improvement in the former case as compared to the latter. We use an initial learning rate of 3e-5 and use early stopping with $patience=5$ to find the best model. For training we use the ICEWS dataset from years 2013-2015 and the year 2016 for testing. The QA generation procedure is described in section~\ref{sec:qa_gen}. There are total 75,788 (37,894$\times$2) examples for training and 5,906 (2,953$\times$2) for test. BERT-RC in Table~\ref{tab:sup_res} indicates the performance of the model on the original test set. We use the RUSS algorithm to obtain simplifications of the test set. BERT-RC-Simple indicates the performance of the model on this simplified test set. It can be observed that simplification brings about an improvement(1.4\%) even on a model that's finetuned on in-domain data. 
\begin{table}[ht!]
\small
\centering
\caption{Table shows the performance of a BERT-base-uncased model finetuned on in-domain dataset. It can be seen that even after finetuning, RUSS approach improves model performance (BERT-RC-Simple).}
\label{tab:sup_res}
\begin{tabular}{@{}ll@{}}
\toprule
\begin{tabular}[c]{@{}l@{}} Model\end{tabular} & F1 \\ \midrule
BERT-RC                                                 & 0.776 \\
BERT-RC-Simple                                          & \textbf{0.787} \\ \bottomrule
\end{tabular}
\end{table}
\section{Related Work} 
Event extraction(EE) has been an active area of research in the past decade. In EE, supervised approaches usually rely on manually labeled training datasets and handcrafted ontologies. ~\citet{li2013joint} utilize the annotated arguments and specific keyword triggers in text to develop an extractor. Supervised approaches have also been studied using dependency parsing by analyzing the event-argument relations and discourse of event interactions~\citep{mcclosky2011event}. These approaches are usually limited by the availability of the fine-grained labeled data and required elaborately designed features.
Recent work formulates event argument extraction as an MRC task. A major challenge with this approach is generating a dataset of QA pairs. ~\citet{liu-etal-2020-event} propose a method combining template based and unsupervised machine translation for question generation. 
~\citet{du-cardie-2020-event} follow a template approach and show that more natural the constructed questions better the event extraction performance. However, none of these methods directly aim to address the long-range dependency problem using simplification.

Automatic text simplification (ATS) systems aim to transform original texts into their lexically and syntactically simpler variants. The motivation for building the first ATS systems was to improve the performance of machine translation systems and other text processing tasks, e.g. parsing, information retrieval, and summarization ~\cite{chandrasekar1996motivations}. In the context of extraction, Zhang et. al. ~\cite{zhang-etal-2018-graph} show that pruning dependency trees to remove irrelevant structures can improve relation extraction performance. Efforts have been made to incorporate syntactic dependencies into models in an effort to mitigate this problem~\citeyear{sha-etal-2016-rbpb,liu-etal-2018-jointly,ma-etal-2020-resource}. Recently, ~\citet{mehta2020simplifythentranslate} have used sentence simplification as a preprocessing step for improving machine translation. Edit-based simplification has been investigated to a great degree to improve the readability of the text~\citep{kumar2020iterative,dong-etal-2019-editnts,alva-manchego-etal-2017-learning}.
To the best of our knowledge this is the first work that studies sentence simplification for improving MRC-based event extraction.

\section{Conclusion \& Future Work}
In this work, we motivated the need for MRC-based event extraction paradigm especially for zero-shot scenarios(\S~\ref{sec:intro}). Next, we discussed the long-range dependency problem ubiquitously faced by event extraction systems. We proposed a context simplification algorithm to reduce the syntactic complexity of the context aided by MRC-system feedback to address the problem(\S~\ref{sec:methods}). Our results indicate that simplification can not only aid MRC systems in a zero-shot setting(\S~\ref{sec:ch3_res}.1-~\ref{sec:ch3_res}.3) but also when they're finetuned on in-domain data(\S~\ref{sec:ch3_res}.4). 

Although, the proposed method can be useful to boost extraction performance when offline computations can be afforded, it may be difficult to scale in real-time use cases. 
Reasons include -- 1) RC system inference time while running the algorithm; 2) call latency if using APIs and 3) search algorithms can be computationally expensive. 
A promising way to improve efficiency would be to generate parallel training data for simplification using the RUSS method offline and guide the MRC model during training using the generated data. This can be done by using attention masks over the deleted words obtained using RUSE. We leave this approach for future work.\\

\noindent{\textbf{Reproducibility}:} Appendix is provided as supplementary material.

\bibliography{anthology,custom}
\bibliographystyle{acl_natbib}

\clearpage
\newpage
\newpage

\appendix

\appendix\section{Appendix}

For training the RUSS algorithm we used the TransformerQA model made available through the allennlp library predictors API~\footnote{\url{https://github.com/allenai/allennlp-models/blob/main/allennlp_models/rc/models/transformer_qa.py}}.
Running the algorithm takes 5 hours on 1 CPU core and 1 GPU. However when parallelizing the computation across 5 cores that time can be brought down to 1 hour.

\section{QA Dataset Generation} \label{sec:qa_gen}
For the ICEWS dataset, we create one question template per predicate per event type for each slot. For each event type we identified a list of most common predicates (triggers) for that event type since trigger labels are not available in the ICEWS dataset. For example, for `Demonstrate or rally' event type the predicates identified are `condemn', `protest', `demonstrate' and for `Accuse' event type the predicates are `blame', `blaming', `accused', `alleged', `accusing'. Table ~\ref{tab:icews_predicates} enumerates all of the ICEWS event types used and their identified predicates. For each of the predicates identified for each event type we use one question template for each of the two argument roles Actor and Target. For the Actor role, the template used an active construction `Who \$predicate\$ someone? and for the same event for the Target role the templated used a passive construction -- `Who was \$predicate\$ by someone?'. This results in 37,894 records with a sentence and two questions one each for the Actor and Target roles respectively. The list of entities for the ICEWS dataset comes from the ICEWS actors and agents dictionaries\footnote{\url{https://dataverse.harvard.edu/dataset.xhtml?persistentId=doi:10.7910/DVN/28118}}.
\begin{table*}[ht!]
\small
\centering
\caption{Table lists the ICEWS event types used and their corresponding predicates that were identified for generating question templates.}
\label{tab:icews_predicates}
\begin{tabular}{lp{8.5cm}}
\toprule
\textbf{Event Type}                                   & \textbf{Predicates}                                                                                                                                           \\ \midrule
Abduct, hijack, or take hostage             & kidnapped, abducting, abducted, captured                                                                                                                          \\
Accuse                                      & blame, blaming, accused, alleged, accusing                                                                                                                        \\
Apologize                                   & apologize, apology                                                                                                                                                \\
Assassinate                                 & carried out assassination of, assassinate                                                                                                                         \\
Bring lawsuit against                       & is suing someone, sued, has sued, filed a suit against                                                                                                            \\
Demonstrate or rally                        & condemn, protest, demonstrate                                                                                                                                     \\
Arrest, detain, or charge with legal action & arrested, sentenced, detained, nabbed, captured, arresting,  capture, jailed, routinely arrested, prosecuted, convicted \\
Use conventional military force             & killed, shelled, combating, shells, strikes, strike, kill    \\                                                               \bottomrule                                    
\end{tabular}

\end{table*}

\section{Dataset Statistics}
Table ~\ref{tab:dataset} outlines the distribution of different event types used in the ICEWS dataset used.
\begin{table*}[ht!]
\centering
\caption{Table shows the distribution of event types in the ICEWS Train and Test datasets used.}
\label{tab:dataset}
\begin{tabular}{@{}lll@{}}
\toprule
Event Type                                  & \#Records Train & \#Records Test \\ \midrule
Abduct, hijack, or take hostage             & 3473            & 193            \\
Accuse                                      & 8856            & 651            \\
Apologize                                   & 181             & 11             \\
Arrest, detain, or charge with legal action & 9933            & 782            \\
Assassinate                                 & 146             & 12             \\
Bring lawsuit against                       & 206             & 18             \\
Demonstrate or rally                        & 2890            & 175            \\
Use conventional military force             & 12209           & 1111           \\ \bottomrule
\end{tabular}
\end{table*}


\end{document}